\documentclass[10pt,twocolumn,letterpaper]{article}

\usepackage{textcomp}
\usepackage{iccv}              

\usepackage{multirow}
\usepackage{graphicx}
\usepackage{bbding}
\usepackage{amsmath}
\usepackage{amssymb}
\usepackage{makecell}
\usepackage{subcaption,booktabs}
\usepackage{xcolor,colortbl}
\definecolor{lavender}{RGB}{150,123,182}
\definecolor{coral}{RGB}{255,127,80}
\definecolor{teal}{RGB}{0,128,128}
\definecolor{lightred}{RGB}{255,204,203}
\definecolor{darkblue}{RGB}{0,51,102}
\usepackage{pifont}
\newcommand{\cmark}{\ding{51}}%
\newcommand{\xmark}{\ding{55}}%

\definecolor{iccvblue}{rgb}{0.21,0.49,0.74}
\usepackage[pagebackref,breaklinks,colorlinks,allcolors=iccvblue]{hyperref}

\usepackage[symbol]{footmisc}

\def\paperID{991} 
\def\confName{ICCV}
\def\confYear{2025}

\newcommand{\methodName}{AdaDrive}

\title{AdaDrive: Self-Adaptive Slow-Fast System for Language-Grounded \\ Autonomous Driving}

\author{ 
    Ruifei Zhang$^{1,2}$, 
    Junlin Xie$^{1,2}$, 
    Wei Zhang$^{4\dagger}$, 
    Weikai Chen$^{\dagger\ddagger}$,
    Xiao Tan$^{4}$, 
    Xiang Wan$^{2}$, 
    Guanbin Li$^{3,5\dagger}$\\
    $^1$ The Chinese University of Hong Kong, Shenzhen 
    $^2$ Shenzhen Research Institute of Big Data \\
    $^3$ Sun Yat-sen University 
    $^4$ Baidu Inc.\\
    $^5$Guangdong Key Laboratory of Big Data Analysis and Processing 
}

\begin{document}
\maketitle

\footnotetext[2]{Co‑corresponding authors: 
\texttt{zhangwei99@baidu.com},
\texttt{chenwk891@gmail.com}, \texttt{liguanbin@mail.sysu.edu.cn}.}
\footnotetext[3]{This paper solely reflects the author's personal research and is not associated with the author's affiliated institution.}

\begin{abstract}
Effectively integrating Large Language Models (LLMs) into autonomous driving requires a balance between leveraging high-level reasoning and maintaining real-time efficiency. Existing approaches either activate LLMs too frequently, causing excessive computational overhead, or use fixed schedules, failing to adapt to dynamic driving conditions. 
To address these challenges, we propose \methodName{}, an adaptively collaborative slow-fast framework that optimally determines when and how LLMs contribute to decision-making. (1) \textbf{When} to activate the LLM: \methodName{} employs a novel adaptive activation loss that dynamically determines LLM invocation based on a comparative learning mechanism, ensuring activation only in complex or critical scenarios. (2) \textbf{How} to integrate LLM assistance: Instead of rigid binary activation, \methodName{} introduces an adaptive fusion strategy that modulates a continuous, scaled LLM influence based on scene complexity and prediction confidence, ensuring seamless collaboration with conventional planners.
Through these strategies, \methodName{} provides a flexible, context-aware framework that maximizes decision accuracy without compromising real-time performance. Extensive experiments on language-grounded autonomous driving benchmarks demonstrate that \methodName{} state-of-the-art performance in terms of both driving accuracy and computational efficiency. Code is available at  \url{https://github.com/ReaFly/AdaDrive}. 

\end{abstract}

\section{Introduction}
\label{sec:intro}
\begin{figure}[t]
\centering
\includegraphics[width=\linewidth, trim=0 0 0 0]{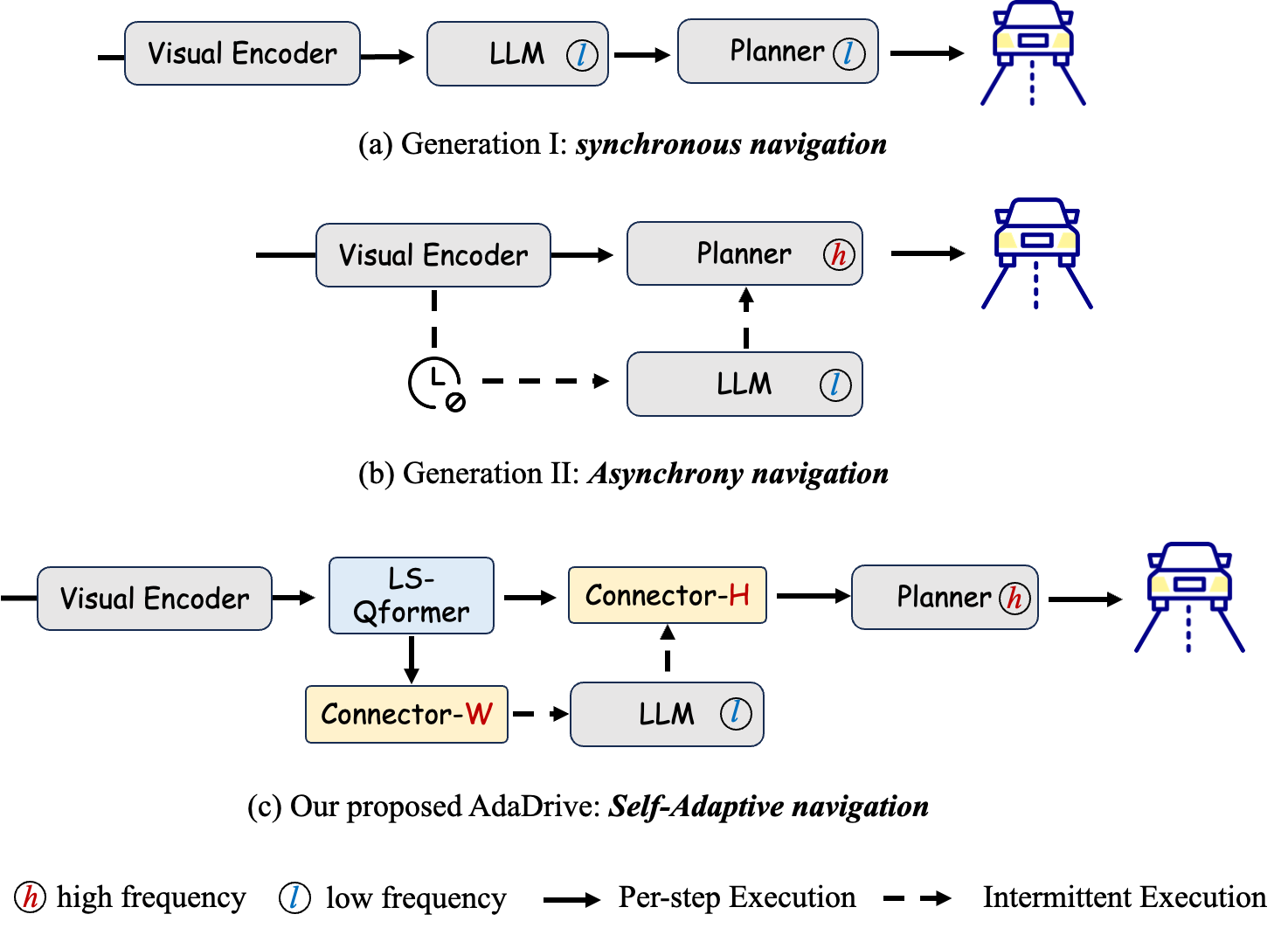}
\caption{(a) The first generation of LLM-enhanced autonomous driving approaches~\cite{shao2023lmdrive,zhang2024ad} employ a synchronous structure, where both LLM and planner operate at each driving step. (b) Generation II methods implement asynchronous processing paradigms, utilizing distinct but predetermined activation frequencies for the LLM and planner. (c) Our proposed AdaDrive also employs an asynchronous architecture but features two novel adaptive connectors: \textbf{Connector-W} for adaptively determining \textbf{when} to activate the LLM, and \textbf{Connector-H} for controlling \textbf{how} to integrate the LLM in driving tasks. This design enables enhanced flexibility in handling uncertain or emergency situations. Besides, we also incorporate LS-Qformer for efficient processing of continuous streaming data.}
\label{fig:first}
\end{figure}

Autonomous driving has long been a focal point in both academia and industry~\cite{codevilla2019exploring,zhang2017query,hawke2020urban,liang2018cirl,toromanoff2020end,zhang2024interactive,chekroun2023gri,zhang2021end,hu2023planning,sadat2020perceive,casas2021mp3,shao2023safety,zhang2024offsetnet}.
With the emergence of large language models (LLMs) and their multimodal extensions (MLLMs), researchers have begun integrating LLMs into autonomous driving systems to enhance cognitive reasoning and decision making~\cite{nie2023reason2drive,sima2023drivelm,mao2023gpt,mao2023language,fu2024drive,cui2024drive,xu2023drivegpt4}.
Early approaches, such as LMDrive~\cite{shao2023lmdrive}
and AD-H~\cite{zhang2024ad}, adopt synchronous and highly-entangled sequential architectures where LLMs continuously influence the driving process at every step (see Figure~\ref{fig:first}). While these models improve driving intelligence, they introduce substantial memory overhead and latency, making real-time deployment challenging, particularly in high-speed, dynamic environments.
To address this issue, subsequent research~\cite{tian2024drivevlm, chen2025asynchronous} has explored asynchronous strategies, where LLM activation occurs at pre-defined intervals to balance performance and efficiency. However, these fixed schedules greatly limit model adaptability, as the need for LLM intervention varies significantly across different driving scenarios. For instances, in safety-critical situations, LLMs may not be invoked when they are needed most. Conversely, in simple scenarios, activating LLMs may be unnecessary, leading to suboptimal resource utilization.

Given these limitations, an ideal LLM-enhanced autonomous driving framework should be able to: 1) \textbf{\textit{Dynamically decide when to activate the LLM}}, ensuring that LLMs contribute only in scenarios where they are beneficial while avoiding unnecessary computational overhead; 2) \textbf{\textit{Adaptively control the degree of LLM influence}}, as our key insight reveals that while LLM engagement consistently enhances performance, a binary on/off activation with full weight (e.g. 1.0) can often be suboptimal compared to a continuous, scaled integration with a lower adaptive weight (e.g. 0.7)  (see results in Table~\ref{tab:components_ablation}: ID~\#3 vs. ID~\#4).

To address these challenges, we introduce \textbf{\methodName{}}, a next-generation self-adaptive LLM-integration framework for autonomous driving. In particular, \methodName{} leverages a slow-fast system paradigm to balance \textit{high-frequency low-latency tasks} (a lightweight planner without invoking the LLM, referred to as, the \textit{fast path}) and \textit{low-frequency high-reasoning tasks} (where the LLM is activated as a cognitive agent, also known as, the \textit{slow path}).

We optimize this slow-fast framework to achieve an optimal balance between decision accuracy and computational efficiency with two key innovations. 
1) \textbf{Adaptive LLM Activations}. Instead of relying on fixed activation intervals, \methodName{} learns when to engage the LLM dynamically through a novel adaptive activation loss. 
By comparing LLM-assisted and LLM-free prediction during training, our model automatically identifies high-risk or complex situations where LLM intervention is most beneficial, ensuring a real on-demand activation.
2) \textbf{Dynamic LLM Contribution Scaling}. Unlike prior methods that treat LLM engagement as a binary decision, \methodName{} introduces a confidence-driven fusion strategy that adjusts the weights of LLM involvement dynamically.
Our key insight is that while LLM assistance consistently improves performance, treating its activation as a binary decision with full weighting can be suboptimal -— adaptive scaling of LLM contributions often yields better results than an all-or-nothing approach.(see results in Table~\ref{tab:components_ablation}: ID~\#3 vs. ID~\#4).
To counter this, \methodName{} modulates the strength of LLM influence based on the confidence of the LLM output and scene complexity, ensuring that its contributions are optimally balanced with conventional planning modules.

In addition, we propose Long-Short Q-former (LS-Qformer) to enhance visual modeling by integrating short-term precision with long-term contextual retention, ensuring consistent trajectory predictions in streaming autonomous driving. We also introduce Propagative Memory Fusion (PMF) mechanism to further optimize memory efficiency by merging evicted frame features into adjacent frames, preserving critical historical context while maintaining a compact representation.
Experimental results demonstrate that \methodName{} sets a new state of art in language-grounded autonomous driving. We summarize our contributions as follows:

\begin{itemize}
    \item We introduce AdaDrive, the first self-adaptive slow-fast architecture for LLM-enhanced autonomous driving, enabling dynamic LLM activation based on real-time driving contexts.
    \item We propose a novel adaptive integration mechanism, which automatically (i) learns when to activate the LLM for maximum performance gains while minimizing computational overhead, and (ii) determines how much the LLM should contribute based on model confidence and scene complexity.
    \item We develop LS-Qformer and PMF mechanism to enhance temporal feature aggregation and preserve critical historical context through efficient memory retention.
    \item We achieve state-of-the-art performance on standard language-grounded autonomous driving benchmarks in terms of both accuracy and computational efficiency.
\end{itemize}

\section{Related Work}
\label{sec:related}

\subsection{End-to-End Autonomous Driving}
Imitation learning~\cite{codevilla2019exploring,zhang2017query,hawke2020urban} and reinforcement learning~\cite{liang2018cirl,toromanoff2020end,chekroun2023gri,zhang2021end} are two primary approaches for end-to-end autonomous driving. Significant advancements have been made in both directions in recent years. 
InterFuser~\cite{shao2023safety} enhances driving safety by effectively leveraging multi-modal, multi-view sensor data and utilizing intermediate interpretable features to constrain actions within a safe set. ReasonNet~\cite{shao2023reasonnet} focuses on global information comprehension and temporal context reasoning, substantially improving the prediction accuracy of object behaviors and enhancing system robustness in challenging scenarios. UniAD~\cite{hu2023planning} proposes a novel modular end-to-end framework that unifies full-stack driving tasks within a single network, enhancing inter-module collaboration for optimal planning performance. 

\begin{figure*}[t]
\centering
\includegraphics[width=\textwidth, trim=0 0 0 0]{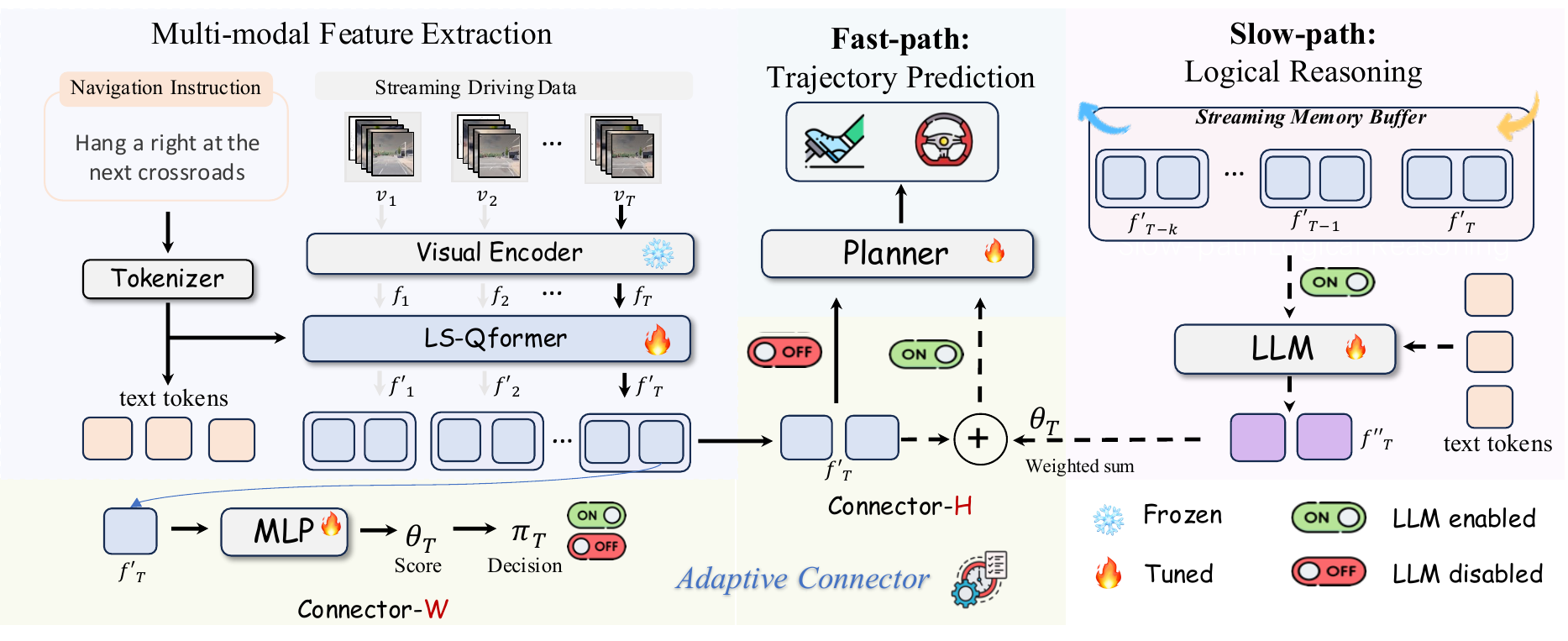}
\caption{An overview of AdaDrive framework, comprising generic multi-modal feature extraction and parallel slow-fast paths dedicated to logical reasoning and trajectory prediction. The two paths are adaptively integrated through our proposed Connector-W and Connector-H components, determining \textbf{when} to activate the LLM and \textbf{how} to integrate the LLM for trajectory prediction, respectively. Dashed lines indicate intermittent execution, which only occurs when LLM is enabled.}
\label{fig:overview}
\end{figure*}

\subsection{LLMs for Autonomous Driving}
Recently, with the emergence of LLMs, their impressive logical reasoning capabilities have catalyzed the integration with autonomous driving systems~\cite{xu2023drivegpt4,shao2023lmdrive,wang2023drivemlm,tian2024drivevlm,chen2025asynchronous,sima2023drivelm}. As a pioneering effort, LMDrive~\cite{shao2023lmdrive} utilizes LLMs to comprehend natural language navigation instructions and predict future waypoints, successfully achieving language-grounded closed-loop autonomous driving. DriveMLM~\cite{wang2023drivemlm} establishes a novel interface between LLMs and autonomous driving systems through semantic mapping of language model reasoning to planners' decision state space.
Despite these advancements, the high computational cost and inference latency associated with LLMs limit their practical applications. To tackle this issue, AsyncDriver~\cite{chen2025asynchronous} presents an asynchronous architecture in which the LLM maintains periodic activations to enhance a traditional planner's capabilities. DriveVLM~\cite{tian2024drivevlm} incorporates the MLLM and traditional planner, where low-frequency activated MLLM provides the reference trajectories to a high-frequency activated planner for trajectory refinement. However, these approaches employ fixed frequencies and invocation intervals for LLMs, severely limiting the collaborative operation of the two systems. In contrast, our framework empowers the planner with autonomous LLM activation capabilities, facilitating dynamic model collaboration while maintaining an optimal balance between task performance and computational resources.

\subsection{MLLMs for Streaming Understanding}
The rapid development of MLLM has showcased versatile capabilities in vision-language comprehension, spatial perception, and video understanding. However, these MLLMs are limited to processing fixed-length images or short clips in an offline manner, constraining their applicability in practical streaming scenarios. Recently, GPT-4o~\cite{openai2024} has demonstrated voice-driven online response capabilities. In parallel, a series of studies ~\cite{zhou2024streaming,chen2024videollm,qian2024streaming,zhang2024flash} have made notable strides in streaming video understanding, further pushing the boundaries of practical applications. VideoLLM-online~\cite{chen2024videollm} pioneers the extension of offline models into online contexts by introducing a novel training objective with a specialized EOS token, prompting the model to remain silent when responses are unnecessary. Flash-VStream~\cite{zhang2024flash} focuses on designing human-like memory modeling to store and process long-term video information while maintaining low inference latency.   
In contrast to video understanding, which focuses on high-level content comprehension and dialogue, autonomous driving in streaming scenarios emphasizes low-level, high-frequency trajectory prediction. This fundamental distinction motivates us to explore a novel paradigm that optimally balances driving performance with inference latency.

\section{Method}
\subsection{Overview}
\paragraph{Problem Definition:}
Given a sequence of streaming video clip data $\mathbf{V}_T=[v_1, v_2, ..., v_T]$ and corresponding navigation instructions $\mathbf{I}_T$,  where $T$ is the current timestamp. This work aims to establish an efficient autonomous driving system $\mathcal{S}$ to generate the instruction-following trajectory prediction:
\begin{equation}
    {W}_T = \mathcal{S}(\mathbf{V}_T,\mathbf{I}_T).
\end{equation}
Here, ${W}_T$ represents the predicted waypoints for timestamp $T$, which is subsequently converted by PID controllers into lateral steering and longitudinal acceleration actions. 

\paragraph{System Architecture:} As shown in Fig.~\ref{fig:overview}, unlike conventional designs where instruction comprehension and trajectory prediction are entangled within LLMs, our proposed \textbf{AdaDrive} decouples these two processes, running them in parallel with distinct activation frequencies. The lightweight planner operates as a low-level trajectory predictor for each frame (\textit{fast path}), while the LLM functions as a central cognitive unit, maintaining low-frequency activation to provide essential assistance to the planner in critical situations (\textit{slow path}). The two paths are adaptively integrated through our proposed Connector-W and Connector-H components, determining when to activate the LLM and how much the LLM should contribute to trajectory prediction, respectively.

\subsection{Slow-Fast Systems}
\paragraph{Multi-modal Feature Extraction:} Given a sequence of streaming
video clip data $\mathbf{V}_T=[v_1, v_2, ..., v_T]$, where each frame data comprises multi-view camera images and point cloud data. We employ a pretrained visual encoder~\cite{shao2023lmdrive} $Vis(\cdot)$ to extract and fuse these multi-modal visual features of each frame: $f_t=Vis(v_t)$, thus constructing $\mathbf{F}_T=[f_1,f_2,...,f_T]$. The subsequent Long-Short Q-former further aggregates the feature tokens in consideration of both long-range and current frame information, denoted as $\mathbf{F}'_T=[f'_1,f'_2,...,f'_T]$, $f'_t \in \mathbb{R}^{N \times C}$, where $N$ is number of tokens and $C$ is the feature dimension. (section \ref{sec:ls-qformer}). 

\paragraph{Fast-path Trajectory Prediction:} The lightweight planner $\mathcal{P}$ maintains high-frequency activation for each timestamp to generate the waypoint only relying on the current frame information: $W_T=\mathcal{P}(f'_T)$. 

\paragraph{Slow-path Logical Reasoning:}
In contrast to the planner, we endow the LLM with access to long-range context information to fully leverage its instruction comprehension and reasoning capabilities.
To prevent unbounded growth in memory usage and computational complexity, making it well-suited for streaming scenarios, we build upon  $\mathbf{F}'_T$ by maintaining a streaming memory buffer to manage the features input for LLM (section \ref{sec:smb}).
This feature buffer maintains a fixed capacity $k$ and we denote the stored features in the buffer as $\mathbf{B}'_T=[f'_{T-k},f'_{T-k+1},...,f'_T]$. Subsequently, the LLM processes the $k$-frame contextual information and outputs the integrated features $f''_T$ for current timestamp $T$ as follows:
\begin{equation}
     f''_T = \mathcal{LLM}(\mathbf I_T, \mathbf{B}'_T)
\end{equation}

\paragraph{Adaptive Connector:} Our framework enhances the slow-fast architecture through adaptive scheduling via two specialized connectors: Connector-W and Connector-H, which orchestrate the interaction between the LLM and the planner. Specifically, Connector-W determines adaptive LLM
activations, while Connector-H controls the dynamic scaling of LLM contributions.

\noindent\textbf{Connector-{\color{red} W}:} Given the current driving context feature $f'_T$ extracted by LS-Qformer, we predict a confidence score that determines the LLM's activation utilizing an MLP function: 
\begin{equation}
    \theta_T=\mathcal{MLP}(f'_T) 
\end{equation}
The continuous probability distribution $\theta_T$ is transformed into a discrete \textbf{binary decision} $\pi_T \in \{0,1\}$ through the Gumbel-Softmax reparameterization, which ensures end-to-end differentiability by maintaining the gradient flow:
\begin{equation}
    \pi_T=\text{Gumbel-Softmax}(\theta_T) 
\end{equation}
However, the optimization of $\pi_T$ presents significant challenges due to the absence of gold standards or ground-truth supervision signals for optimal activation timing. In our work, we propose a novel comparative learning based adaptive activation loss, to address these issues. Specifically, in the training stage, We perform two forward passes for trajectory prediction: one with LLM assistance yielding $W_T^{LLM}=\mathcal{P}(f'_T+f''_T)$, and another without, producing $W_T=\mathcal{P}(f'_T)$. Subsequently, we calculate the trajectory loss (L1 loss) for $W_T^{LLM}$ and $W_T$, denoted as $\mathcal{L}_T^{LLM}$ and $\mathcal{L}_{T}$, respectively. Following a warmup phase where both losses converge to stable values, their comparative difference reflects the magnitude of LLM's contribution to trajectory prediction at the current timestep. Thus, we link the binary decision $\pi_T$ with the trajectory losses to construct a novel \textbf{adaptive activation loss}:
\begin{equation}
    \mathcal{L}_{ada} = \pi_T * \mathcal{L}_T^{LLM} + (1-\pi_T)* \mathcal{L}_{T}
\end{equation}
Optimizing this objective function naturally induces $\pi_T = 1$ when $\mathcal{L}_T^{LLM} < \mathcal{L}_{T}$ and $0$ otherwise, thereby enabling the model to learn optimal LLM activation conditions.
Further, to achieve optimal performance gains while minimizing computational overhead, we introduce a penalty term $\gamma$ into the LLM-assisted trajectory loss to control the frequency of LLM activations, ensuring LLM is only activated when $\mathcal L_T^{LLM}$ is significantly lower than $\mathcal L_T$ by a predetermined margin $d$:
\begin{equation}
    \mathcal{L}_{ada} = \pi_T * (\mathcal L_T^{LLM} + \gamma) + (1-\pi_T)* \mathcal L_T
\end{equation}
\begin{equation}
    \gamma = max(d - (L_T- L_T^{LLM}),0)
\end{equation}

\begin{figure}[t]
\centering
\includegraphics[width=0.95\linewidth, trim=0 0 0 0]{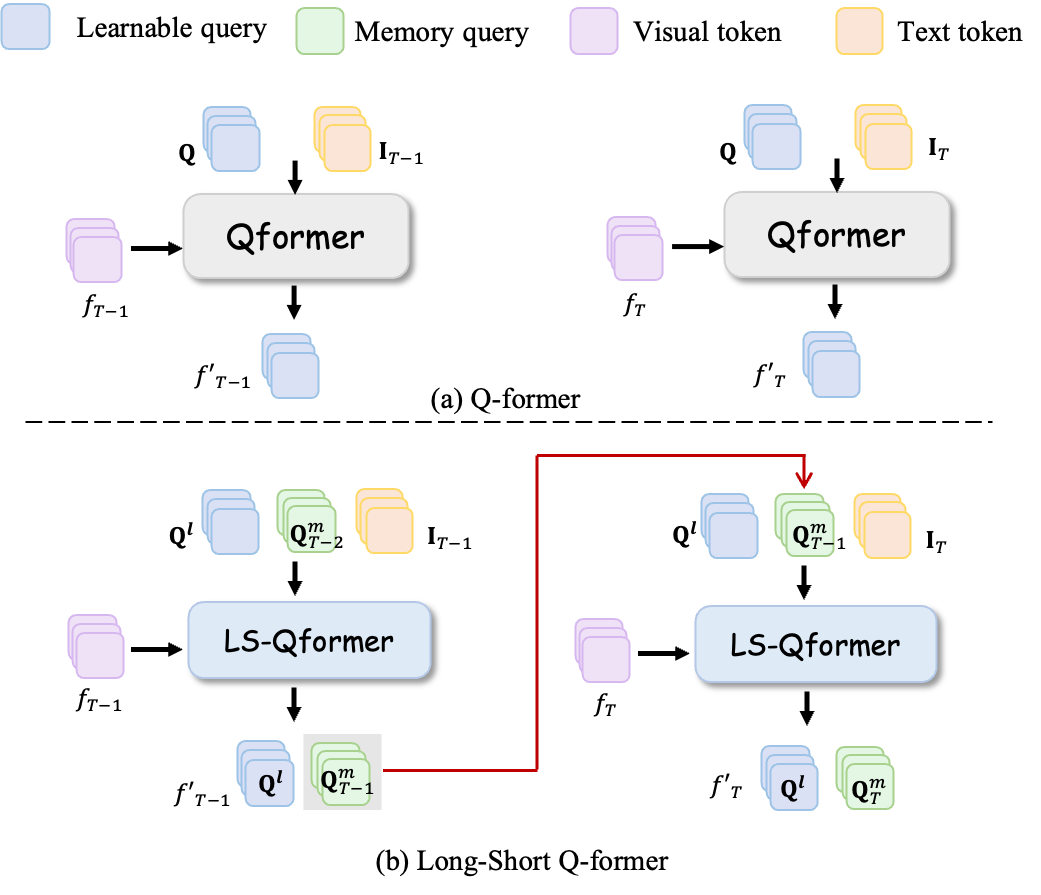}
\caption{Comparisons between the Q-former and our proposed Long-Short Q-former (LS-Qformer). }
\label{fig:lsqformer}
\end{figure}

\noindent\textbf{Connector-{\color{red}H}:} 
Through our proposed adaptive activation loss, the model (Connector-W) learns to determine optimal LLM activation timing. However, binary fusion (all-or-nothing) may not be the optimal strategy for seamless integration with conventional planners. To enable dynamic LLM contribution scaling, Connector-H leverages the predicted confidence score $\theta_T$ as a fusion coefficient for weighted feature integration, generating a third trajectory prediction $W_T^{Fuse}=\mathcal{P}({f}'_T + \theta_T * {f}''_T)$. The trajectory loss computed for $W_T^{Fuse}$ inherently guides the model to learn optimal contribution scaling.

\paragraph{Inference Stage:} Connector-W predicts the confidence score $\theta_T$ and corresponding binary decision $\pi_T$ for LLM's activation. Upon LLM activation, Connector-H modulates the contribution of LLM features $f''_T$ to base features $f'_T$ by leveraging the prediction confidence $\theta_T$ as a dynamic weighting coefficient. Specifically, the trajectory prediction can be uniformly formulated as follows:
\begin{equation}
    {W}_T = 
    \begin{cases}
        \mathcal{P}({f}'_T), & \text{if LLM is not activated ,} \\
        \mathcal{P}({f}'_T + \theta_T * {f}''_T) , & \text{if LLM is activated.}
    \end{cases}
\end{equation}

\subsection{Long-Short Feature Modeling}
\label{sec:ls-qformer}
As a common connector to bridge visual encoder and LLM, Q-former has been applied in many MLLMs. The vanilla Q-former can be formulated as follows:
\begin{equation}
    f'_T = \text{Q-former}(\mathbf{Q}, f_T, \mathbf{I}_T)
\end{equation}
where $\mathbf{Q}$ is additional introduced learnable tokens for feature aggregation. However, this module processes each frame separately while ignoring the long-range temporary information. To tackle this issue, we propose a Long-Short Q-former. Inspired by the group mechanism~\cite{cohen2016group}, we partition the learnable tokens into two groups, denoted as $\mathbf{Q}^m$ and $\mathbf{Q}^l$. $\mathbf{Q}^m$ is propagated into the next frame for aggregating long-range information while $\mathbf{Q}^l$ are similar to the vanilla Q-former focusing on the current frame:
\begin{equation}
    {f}'_T= [\mathbf{Q}^l; \mathbf{Q}^m_T] = \text{Q-former}(\mathbf{Q}^l, \mathbf{Q}^m_{T-1}, {f}_T, \mathbf{I}_T)
\end{equation}
Through this mechanism, LS-Qformer simultaneously extracts critical features from current frames and models temporal feature evolution, yielding richer visual representations.

\begin{figure}[t]
\centering
\includegraphics[width=\linewidth, trim=0 0 0 0]{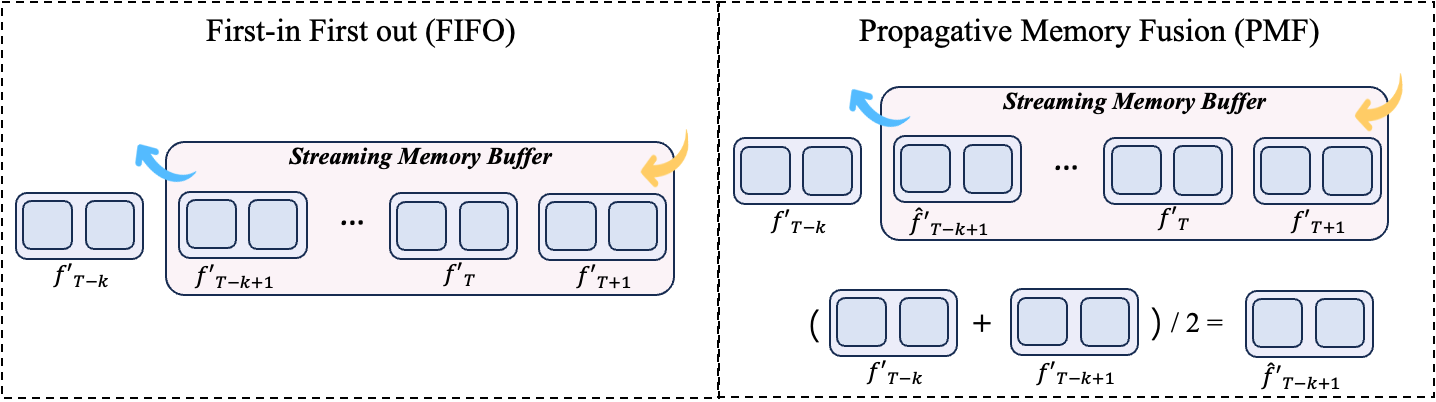}
\caption{Illustration of FIFO and our proposed PMF. Unlike FIFO, PMF maintains a compact buffer while enabling forward information flow by merging features from to-be-evicted frames into their preceding frames.}
\label{fig:smb}
\end{figure}

\subsection{Streaming Memory Buffer}
\label{sec:smb}
Long-range contextual information is crucial for predicting objects' potential behaviors and trajectories, thereby enabling safer autonomous driving. However, storing and processing continuous streaming data inevitably leads to exponential growth in computational overhead and potential memory overflow. To address these challenges, we propose a fixed-size streaming memory buffer with a Propagative Memory Fusion (PMF) strategy for managing historical driving data (illustrated in Fig.~\ref{fig:smb}). Compared to First-in-First-out (FIFO) which only retains fixed-length features, our PMF mechanism preserves information by merging features of the to-be-evicted frame into its preceding frame, maintaining a compact buffer while enabling forward information propagation: 
\begin{equation}
    \hat{f}'_{T-k+1} = (f'_{T-k} + f'_{T-k+1}) / 2
\end{equation}
Subsequently, the memory buffer is updated to $\mathbf{B}'_T=[\hat {f}'_{T-k+1},f'_{T-k+2},...,f'_{T+1}]$, where $\hat {f}'_{T-k+1}$ represents the fused features.

\section{Experiments}
\begin{table*}[t]
\centering
\caption{Performance comparison of our method with state-of-the-art approaches on the LangAuto-Tiny and LangAuto-Short benchmarks.}
\resizebox{0.98\textwidth}{!}{
\renewcommand{\arraystretch}{1.1}
\begin{tabular}{cccccccc}
\toprule
\multirow{2}{*}{Method}  & \multirow{2}{*}{LLM (\#Params)} & \multicolumn{3}{c}{LangAuto-Tiny} & \multicolumn{3}{c}{LangAuto-Short} \\ 
\cmidrule(r){3-5} \cmidrule(r){6-8} 
                       & &  DS $\uparrow$  & RC $\uparrow$   &  IS $\uparrow$      &  DS $\uparrow$  & RC $\uparrow$   &  IS $\uparrow$   \\ \cmidrule(r){1-2} \cmidrule(r){3-5} \cmidrule(r){6-8} 
\multirow{4}{*}{LMDrive~\cite{shao2023lmdrive}} 
                  &  LLaMA2 (7B)~\cite{touvron2023llama2}          &  56.1    &  64.2    &  0.87 &  44.8  &  53.5    &  0.84 \\ 
                  &  Vicuna-v1.5 (7B)~\cite{zheng2023judging}   &  59.0    &  69.9    &  0.84 &   47.0   &  56.5    &  0.83 \\ 
                  &  LLaVA-v1.5 (7B)~\cite{liu2023improved}      &  66.5    &  77.9   &  0.85 &   50.6   &  60.0    &  0.84   \\ 
                  &   TinyLLaMA (1.1B)~\cite{zhang2024tinyllama}     &   {64.1}  &  {75.0}  &  {0.86 }  &  {46.2} &  {59.7}  &  {0.79}   \\ 
                  \midrule
AD-H~\cite{zhang2024ad}  & Mipha (3B)~\cite{zhu2024comprehensive} + OPT (350M)~\cite{zhang2023opt} & 68.0 &  74.4 & 0.87 & 54.3   &  61.8  & \textbf{0.86}  \\
\midrule
AdaDrive & TinyLLaMA (1.1B) ~\cite{zhang2024tinyllama} + Planner (3M)& \textbf{80.9} & \textbf{87.6} & \textbf{0.90} &\textbf{70.6}   &  \textbf{85.3}  & 0.81  \\
\bottomrule
\end{tabular}}

\label{table:compare_tiny_short}
\end{table*}

\begin{table*}[t]
\centering
\caption{Performance comparison of our method with state-of-the-art approaches on the LangAuto benchmark.}
\resizebox{0.98\textwidth}{!}{
\renewcommand{\arraystretch}{1.1}
\begin{tabular}{ccccccc}
\toprule
\multirow{2}{*}{Method}  &  \multirow{2}{*}{LLM (\#Params)} & \multicolumn{3}{c}{LangAuto} & \multirow{2}{*}{Mem $\downarrow$} & \multirow{2}{*}{Inf. Time $\downarrow$}\\ \cmidrule(r){3-5} 
                       & & DS $\uparrow$  & RC $\uparrow$   &  IS $\uparrow$  & (G)  & (ms)  \\ \cmidrule(r){1-2}  \cmidrule(r){3-5} \cmidrule(r){6-7} 
\multirow{4}{*}{LMDrive~\cite{shao2023lmdrive}}  &   
                  LLaMA2 (7B) ~\cite{touvron2023llama2}   & 32.8   &  40.1  & 0.81  & \multirow{3}{*}{26.91}     & \multirow{3}{*}{526}   \\ 
                  
                  &  Vicuna-v1.5 (7B) ~\cite{zheng2023judging}  &  34.0  &  39.0  &  0.85  &     &  \\ 
                  &   LLaVA-v1.5 (7B) ~\cite{liu2023improved}   &  36.2  &  46.5 &  0.81  &  &    \\ 
                   
                  & TinyLLaMA (1.1B) ~\cite{zhang2024tinyllama} &  25.2 &  38.6  & 0.71 & 16.29  &445   \\ 
                  \midrule
AD-H~\cite{zhang2024ad}  & Mipha (3B)~\cite{zhu2024comprehensive} + OPT (350M)~\cite{zhang2023opt} & 41.1 &  48.5 & \textbf{0.86} & -   &  -    \\
\midrule
AdaDrive   & TinyLLaMA (1.1B) ~\cite{zhang2024tinyllama} + Planner (3M) & \textbf{42.9} &  \textbf{53.4} & 0.82 &  \textbf{6.79}  & \textbf{189 }    \\
\bottomrule                    
\end{tabular}}
\label{table:compare_long}
\end{table*}

\begin{table*}[t]
\centering
\begin{minipage}{.48\textwidth}
    \centering
    \caption{Ablation of Connector and LS-Qformer components on the LangAuto-Tiny benchmark.}
    \vspace{-0.05in}
    \label{tab:components_ablation}
    \resizebox{0.95\linewidth}{!}{
        \renewcommand{\arraystretch}{1.1}
        \begin{tabular}{>{\kern-0.5\tabcolsep}c|ccc|ccc<{\kern-0.5\tabcolsep}}
            \toprule
            \multirow{2}{*}{ID} &\multicolumn{2}{c}{Connector} & \multirow{2}{*}{LS-Qformer} &  \multirow{2}{*}{DS $\uparrow$} & \multirow{2}{*}{RC$\uparrow$} & \multirow{2}{*}{IS$\uparrow$} \\
            \Xcline{2-3}{0.3pt}
            & W & H & & &  & \\
            \midrule
             1&\xmark & \xmark & \xmark & 67.4 & 75.3 & 0.86 \\
             2&\xmark & \xmark & \cmark& 71.9 & 82.6 & 0.84 \\
             3& \cmark &\xmark & \cmark & 77.9 & 84.8 & 0.89 \\
             4&\cmark & \cmark&\cmark &  \textbf{80.9 }&  \textbf{87.6} &  \textbf{0.90} \\
            \bottomrule
        \end{tabular}
    }
\end{minipage}
\hfill
\begin{minipage}{.48\textwidth}
    \caption{Ablation of different feature modeling methods on the LangAuto-Tiny benchmark.}
    \label{tab:temmporal_ablation}
    \vspace{-0.05in}
    \resizebox{0.99\linewidth}{!}{
        \renewcommand{\arraystretch}{1.1}
        \begin{tabular}{>{\kern-0.5\tabcolsep}c|c|ccc<{\kern-0.5\tabcolsep}}
            \toprule
             Method  &  \#Token & DS $\uparrow$& RC$\uparrow$ &  IS $\uparrow$ \\
            \midrule
             Q-former~\cite{li2023blip} & 40   &75.8 & 83.4   &  0.88\\
             SeqQ-Former~\cite{ma2023vista} &40  &77.6     & 83.5 & 0.89\\
             Q-former\&Add & 40  &77.4  &83.5  & 0.89 \\
             LS-Qformer (Ours) & 20+20 & \textbf{80.9} & \textbf{ 87.6} & \textbf{0.90} \\
            \bottomrule
        \end{tabular}
    }
\end{minipage}
\end{table*}

\subsection{Experimental Setup}
\paragraph{Dataset:} 
We train the AdaDrive on the standard LangAuto dataset~\cite{shao2023lmdrive}, a comprehensive multi-modal collection comprising 64K instruction-following sequences. Each sequence encapsulates synchronized multi-view camera images and LiDAR point clouds, providing rich spatiotemporal context for autonomous navigation. 
\paragraph{Benchmarks and Metrics:} We conduct closed-loop autonomous driving evaluations on the LangAuto benchmark within the CARLA simulation environment, where the benchmark is structured into three distinct subtasks based on driving distances: LangAuto-Tiny, LangAuto-Short, and LangAuto. Route completion (RC), infraction score (IS), and driving score (DS) are three widely adopted evaluation metrics. Specifically, RC denotes the ratio of successfully traversed distance by an agent to the total planned route length. IS aggregates multiple categories of traffic violations through geometric progression, initializing at 1.0 and decaying multiplicatively with each infraction occurrence. DS synthesizes route completion and infraction penalties through multiplication, serving as the principal evaluation criterion and providing a comprehensive assessment of autonomous driving performance.
\paragraph{Model Configuration:} Our framework employs a pretrained visual encoder from \cite{shao2023lmdrive}, which remains frozen during training. For language modeling, we adopt TinyLLaMA~\cite{zhang2024tinyllama}, a lightweight language model, to reduce computational overhead and parameter count. The planner adopts a 4-layer Transformer~\cite{vaswani2017attention} architecture. We adopt 20 learnable tokens and 20 memory tokens in LS-Qformer and set the capacity $k$ of streaming memory buffer to 10.
\paragraph{Implementation Details:}  
We employ an AdamW optimizer with a cosine learning rate scheduler. The initial learning rate is set to $1\times10^{-5}$ with training spanning 15 epochs. In the loss function, we set the hyper-parameter margin $d$ to 0.3 to constrain the LLM activation.

\begin{figure*}[t]
\centering
\includegraphics[width=\linewidth, trim=0 0 0 0]{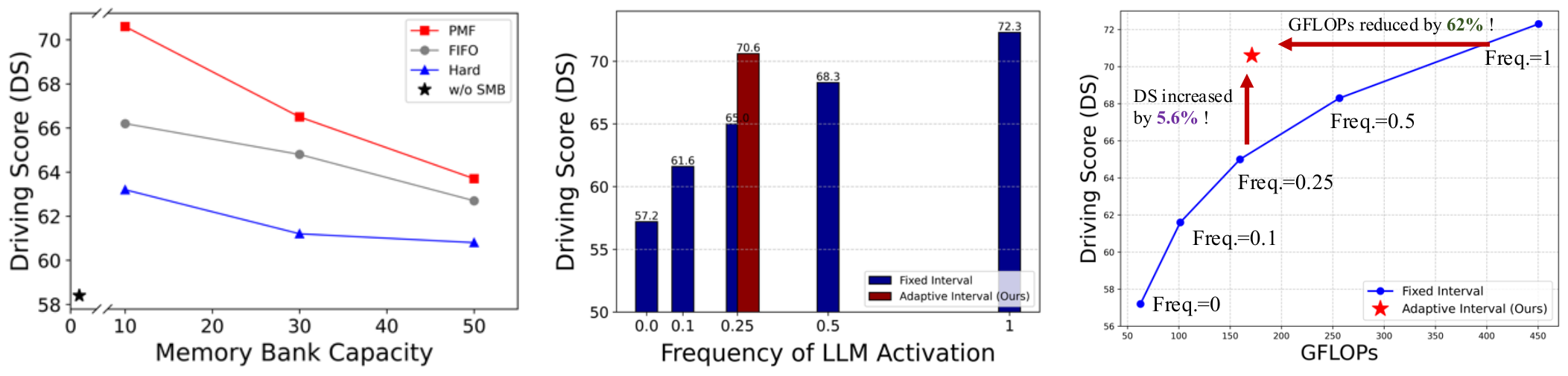}
\vspace{-0.25in}
\caption{(a) Ablation on varying streaming memory buffer (SMB) capacities and content update mechanisms. (b) Comparison of our self-adaptive LLM activation \textit{vs.} fixed-interval activation (freq. = 0, 0.1, 0.25, 0.5, and 1, where 0 indicates no activation and 1 indicates full activation) on driving scores. (c) Comparison of our self-adaptive LLM activation with fixed-interval LLM activation in terms of computational cost (GFLOPs) and driving scores. These analyses are performed using the LangAuto-Short benchmark.}
\vspace{-0.1in}
\label{fig:mbsize_llm}
\end{figure*}

\subsection{Main Results}
\paragraph{Closed-loop Driving Performance:} 
We conduct comprehensive experiments to evaluate our method on the LangAuto benchmarks~\cite{shao2023lmdrive}, comparing against state-of-the-art approaches including LMDrive~\cite{shao2023lmdrive} and AD-H~\cite{zhang2024ad}. It is worth noting that AD-H employs additional mid-level language commands to train its hierarchical multi-agent driving system. The experimental results are presented in Tables~\ref{table:compare_tiny_short} and~\ref{table:compare_long}. Our proposed AdaDrive demonstrates superior performance across all distance-based sub-tracks, particularly excelling in tiny and short route scenarios. Specifically, AdaDrive achieves driving scores of 80.9\% and 70.6\% on the LangAuto-Tiny and LangAuto-Short benchmarks, surpassing the second-best method AD-H by significant margins of 12.9\% and 16.3\%, respectively. These results validate the effectiveness of our self-adaptive slow-fast driving system.
\paragraph{Inference Time and Memory Cost:} In addition to enhanced driving performance, our method exhibits substantial advantages in inference time and computational overhead, as showcased in Table~\ref{table:compare_long}. These benefits are attributed to two key architectural designs: 1) the self-adaptive slow-fast system. Unlike LMDrive and AD-H, which adopt sequential processing requiring LLM inference at every timestamp, our parallel architecture primarily relies on a lightweight planner, with the LLM activated only during emergencies as determined by the system's adaptive scheduling. Moreover, the planner only needs to process the current frame features, as the historical information has been propagated through the LS-Qformer. These architectural designs significantly reduce the system's inference latency. 2) The tailored streaming memory buffer. Existing methods lack specialized handling for streaming inputs, leading to data accumulation and increased memory overhead. In contrast, we explicitly propose a streaming memory buffer architecture that efficiently manages input data, reducing memory costs while improving inference speed.

\subsection{Ablation Study}
\paragraph{Components Effectiveness:}
We conduct comprehensive ablation studies to validate the effectiveness of the proposed LS-Qformer and quantify the performance gains achieved through connector-driven LLM interaction. First, we start from the baseline which implements a vanilla Q-former that independently aggregates frame-level features and a planner for trajectory prediction (ID~\#1). The results are presented in Table~\ref{tab:components_ablation}. Replacing the vanilla Q-former with our proposed LS-Qformer (ID~\#2) yields substantial performance improvements. These results demonstrate that the LS-Qformer effectively captures temporal dependencies in historical information, enabling more informed planning decisions. Furthermore, we integrate the LLM into our system through the Connector architecture. Leveraging the dynamic LLM activation mechanism governed by Connector-W (ID~\#3), our approach achieves significant performance gains, attaining a driving score of 77.9\%. Moreover, by replacing the conventional full weighting LLM's feature fusion with our innovative Connector-H-controlled dynamic LLM contribution scaling strategy (ID~\#4), we observe further enhancement in the overall DS performance metrics.

\paragraph{Analysis of LS-Qformer:}
We compare our LS-Qformer against several architectural variants: 1) the standard Q-former which processes frames independently in a frame-wise manner; 2) SeqQ-Former~\cite{ma2023vista}, which propagates current output tokens as queries for subsequent frame feature extraction; and 3) Q-former with temporal accumulation, which incorporates historical context by additively fusing token representations from previous frames with current frame features. The results in Table~\ref{tab:temmporal_ablation} demonstrate that our LS-former achieves optimal driving scores by ingeniously incorporating long-range historical information with current frame content through a grouping mechanism.

\paragraph{Analysis of Streaming Memory Buffer:} We investigate the impact of different memory bank capacities and content update mechanisms on trajectory prediction, as illustrated in Fig.~\ref{fig:mbsize_llm}(a). Several key observations emerge: 1) Thanks to our LS-Qformer's effective context aggregation, our method achieves comparable performance even when the LLM only attends to the current frame (w/o SMB). 2) Smaller memory bank capacities prove more beneficial for trajectory prediction. We hypothesize that as memory content increases, the LLM's instruction perception capability becomes diluted among the expanded context. 3) The hard update mechanism, which completely clears the current buffer upon reaching capacity limits, introduces inherent instabilities in subsequent trajectory predictions. In contrast, the PMF mechanism maintains temporal coherence while preserving more contextual information, leading to superior performance.

\begin{figure}[t]
\centering
\includegraphics[width=0.98\linewidth, trim=0 0 0 0]{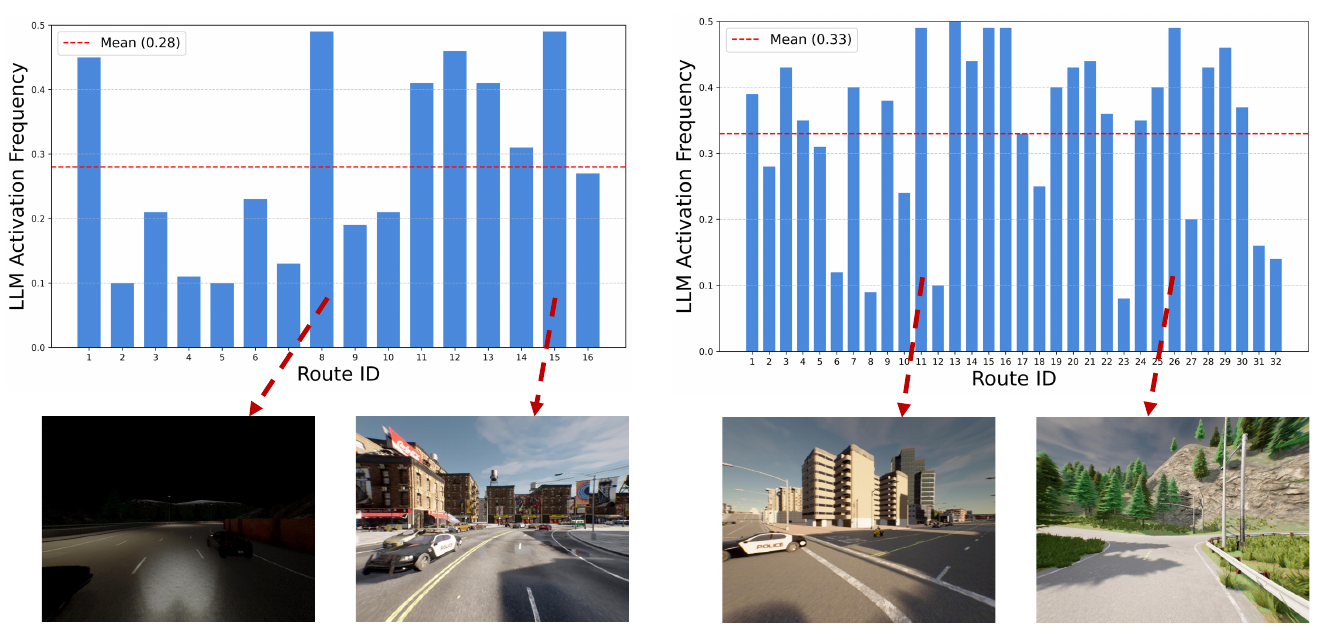}
\caption{The distribution of LLM activation frequencies across all routes in both LangAuto-Short and LangAuto benchmarks.}
\label{fig:routellm}
\vspace{-3mm}
\end{figure}

\paragraph{Analysis of Adaptive Collaboration:}
We compare our adaptive LLM activation strategy against fixed-interval activation at various frequencies. As illustrated in Fig~\ref{fig:mbsize_llm}(b), higher activation frequencies consistently yield more stable and robust driving performance. Our adaptive LLM activation mechanism enables dynamic responses to critical scenarios, achieving comparable performance to continuous LLM activation (frequency = 1.0) while maintaining an average activation frequency of only \textbf{0.28}. Fig~\ref{fig:mbsize_llm}(c) further demonstrates that our method strikes an optimal balance between driving performance and computational efficiency, reducing GFLOPs by 62\% compared to continuous activation (frequency = 1.0) while improving driving scores by 5.6\% relative to the fixed-interval scheme with a similar frequency (frequency = 0.25).

\begin{figure}[t]
\centering
\includegraphics[width=0.98\linewidth, trim=0 0 0 0]{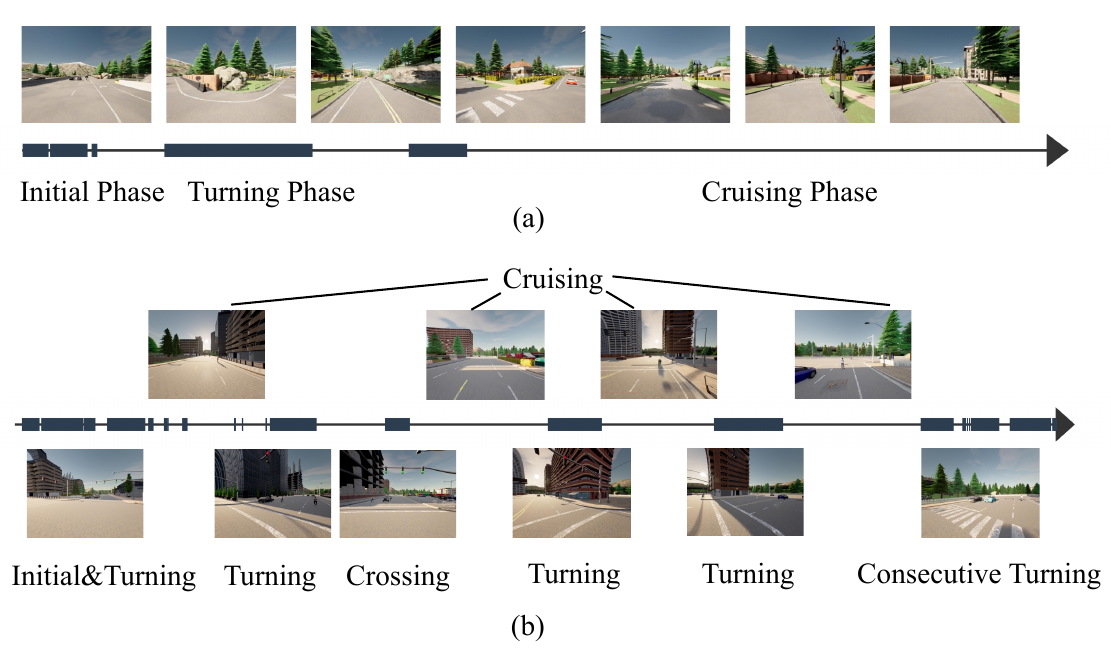}
\vspace{-0.1in}
\caption{Temporal distribution of LLM activations for Route 7 (a) and Route 9 (b) in the LangAuto-Short benchmark. LLM activation moments, highlighted in {\color{darkblue}{darkblue}} on the timeline, demonstrate concentrated engagement during complex maneuvers such as turning and crossing, while remaining dormant during routine cruising phases.}
\label{fig:routellm-single}
\vspace{-5mm}
\end{figure}

Besides, we analyze the distribution of LLM activation frequencies across all routes in both LangAuto-Short and LangAuto benchmarks, as illustrated in Fig.~\ref{fig:routellm}. The activation frequencies range from 0.1 to 0.5, demonstrating effective sparsity and dynamic adaptation, with average activation rates of 0.28 and 0.33 respectively. Notably, higher activation frequencies are observed in challenging routes, such as dense urban streets, nighttime conditions or mountain roads, validating our design principle of adaptive LLM engagement for complex situations. 
Furthermore, by analyzing the temporal distribution of LLM activations within individual routes, we identify patterns of increased LLM engagement during critical driving steps. As illustrated in Fig.~\ref{fig:routellm-single}, LLM activations are predominantly concentrated in complicated scenarios, including directional transitions, and intersection navigation. The LLM's advanced logical reasoning capabilities significantly enhance the autonomous vehicle agent's decision-making performance in these situations.

\section{Conclusion}
This work explores LLM-powered language-grounded autonomous driving, focusing on two fundamental questions: optimal activation timing and effective utilization strategies of LLMs. Specifically, our approach features a self-adaptive slow-fast architecture that adaptively schedules LLM activation according to driving situations, while dynamically modulating its contribution weight based on prediction confidence scores. This strategy significantly enhances model flexibility and robustness while maintaining controlled computational overhead. Additionally, we introduce a tailored LS-Qformer for effective historical context aggregation and a streaming memory buffer with a propagative memory fusion strategy for efficient unbounded temporal data management. Extensive experiments demonstrate that our approach significantly outperforms existing methods in both effectiveness and efficiency, validating its potential for practical applications.

\section*{Acknowledgments}
This work is supported in part by the National Key R\&D Program of China (2024YFB3908503), in part by the National Natural Science Foundation of China (62322608), in part by the Shenzhen Longgang District Science and Technology Innovation Special Fund (No. LGKCYLWS2023018), in part by the Futian Healthcare Research Project (No.FTWS002), and in part by the Shenzhen Medical Research Fund (No. C2401036). This work is also sponsored by CIE-Tencent Robotics X Rhino-Bird Focused Research Program. 

{
    \small
    \bibliographystyle{ieeenat_fullname}
    \bibliography{main}
}

\end{document}